\RequirePackage{amsmath}
\documentclass[runningheads]{llncs}
\usepackage{graphicx}
\usepackage{siunitx}
\usepackage{hyperref}
\usepackage{amsmath,amssymb,amsfonts}
\usepackage{algorithmic}
\usepackage{booktabs}
\usepackage{tikz-cd}
\usetikzlibrary{shapes}
\usetikzlibrary{arrows}
\usepackage[linesnumbered,algoruled,boxed,lined,ruled]{algorithm2e}
\usepackage{textcomp}
\usepackage{xcolor}

\makeatletter
\newcommand\@erelb@r[1]{%
  \mathrel{\tikz[baseline=-.5ex]\draw[#1] (0,0)--(.5,0);}
}
\newcommand{\erelbar}[1]{\@erelbar#1}
\def\@erelbar#1#2{%
  \ifcase\numexpr#1*4+#2\relax
    \@erelb@r{-}\or     
    \@erelb@r{->}\or    
    \@erelb@r{-|}\or    
    \@erelb@r{-o}\or   
    \@erelb@r{<-}\or    
    \@erelb@r{<->}\or   
    \@erelb@r{<-|}\or   
    \@erelb@r{<-o}\or   
    \@erelb@r{|-}\or    
    \@erelb@r{|->}\or   
    \@erelb@r{|-|}\or   
    \@erelb@r{|-o}\or 
    \@erelb@r{o-}\or   
    \@erelb@r{o->}\or  
    \@erelb@r{o-|}\or  
    \@erelb@r{o-o}    
  \else
    \@wrong
  \fi
}

\begin{document}
\title{Order-Independent Structure Learning of Multivariate Regression Chain Graphs\thanks{Supported by AFRL and DARPA (FA8750-16-2-0042).}}
\titlerunning{Order-Independent Structure Learning of MVR CGS}
%
\author{Mohammad Ali Javidian \and Marco Valtorta \and Pooyan Jamshidi}

\authorrunning{Javidian et al.}
%
\institute{University of South Carolina \\
\email{javidian@email.sc.edu, mgv@cse.sc.edu, pjamshid@cse.sc.edu}}
 \maketitle              
\begin{abstract}This paper deals with multivariate regression chain graphs
	(MVR CGs), which were introduced by Cox and Wermuth \cite{cw1,cw2} to represent linear causal models with correlated errors. We consider the PC-like
	algorithm for structure learning of MVR CGs, which is a constraint-based
	method proposed by Sonntag and Pe\~{n}a in \cite{sp}. We show that the PC-like
	algorithm is order-dependent, in the sense that the output can depend
	on the order in which the variables are given. This order-dependence is
	a minor issue in low-dimensional settings. However, it can be very pronounced in high-dimensional settings, where it can lead to highly variable
	results. We propose two modifications of the PC-like algorithm that remove part or all of this order-dependence. Simulations under a variety
	of settings demonstrate the competitive performance of our algorithms
	in comparison with the original PC-like algorithm in low-dimensional
	settings and improved performance in high-dimensional settings.

\end{abstract}
\section{Introduction}
Chain graphs were introduced by Lauritzen, Wermuth and Frydenberg \cite{f},\cite{lw} as a generalization of graphs based on undirected graphs and directed acyclic graphs (DAGs). Later Andersson, Madigan 
and Perlman introduced an alternative Markov property for chain graphs \cite{amp}.
In 1993 \cite{cw1}, Cox and Wermuth introduced multivariate regression chain graphs (MVR CGs). The different interpretations of CGs
have different merits, but none of the interpretations subsumes another interpretation \cite{d}.

Acyclic directed mixed graphs (ADMGs), also known as semi-Markov(ian) \cite{pj} 
models contain directed ($\rightarrow$) and bidirected
($\leftrightarrow$) edges subject to the restriction that there are no directed cycles \cite{r2}.
An ADMG that has no partially directed cycle is called a \textit {multivariate regression chain graph}. Cox and Wermuth represented these graphs
using directed edges and dashed edges, but we follow Richardson \cite{r2} because bidirected edges allow the \textit{m}-separation criterion (defined in section \ref{definitions})
to be viewed more directly as an extension of $d$-separation than is possible with dashed edges \cite{r2}.


Unlike in the other CG interpretations, the bidirected edge in MVR CGs has
a strong intuitive meaning. It can be seen to represent one or more hidden
common causes between the variables connected by it. In other words, in an MVR CG any bidirected
edge $X\leftrightarrow Y$ can be replaced by $X\gets H\to Y$ to obtain a Bayesian network representing
the same independence model over the original variables, i.e. excluding the
new variables H. These variables are called hidden, or latent, and have been
marginalized away in the CG model.
See \cite{sonntagpena15},\cite{jv1}, \cite{essentialmvrcgs} for details on the properties of MVR chain graphs. 

Two \textit{constraint-based} learning algorithms, that use a statistical analysis to test the presence of a
conditional independency, exist for learning MVR CGs:
(1) the PC-like algorithm \cite{sp}, and (2) the answer set programming (ASP) algorithm \cite{Pena2016}. The PC-like algorithm extends the original learning algorithm for Bayesian networks by \textbf{P}eter Spirtes and \textbf{C}lark Glymour~\cite{sgs}. It learns the structure of the underlying MVR chain graph in four steps: (a) determining the skeleton: the resulting undirected graph in this phase contains an undirected edge $u-v$ iff there is no set $S\subseteq V\setminus\{u,v\}$ such that $u\!\perp\!\!\!\perp v|S$; (b) determining the \textit{v}-structures (unshielded colliders); (c) orienting some of the undirected/directed edges into directed/bidirected edges according to a set of rules applied iteratively; (d) transforming the resulting graph in the previous step into an MVR CG. The essential recovery algorithm obtained after step (c) contains all directed and bidirected edges that are present in every MVR CG of the same Markov equivalence class.

In this paper, we show that the PC-like algorithm is order-dependent,
in the sense that the output can depend on the order in which the variables are given. We propose several modifications of the PC-like algorithm that remove part or all of this order-dependence, but do not change the result when perfect conditional independence information is used. When
applied to data, the modified algorithms are partly or fully order-independent. Details of  experimental results can be found in the supplementary material at \url{https://github.com/majavid/SUM2019}.

 Our main contributions are the following:
 \begin{enumerate}
     \item We propose several modifications of the PC-like algorithm for learning the structure of MVR chain graphs under the faithfulness assumption that remove part or all of the order-dependence.
     \item We compared the performance of our algorithms with that of the PC-like algorithm proposed in \cite{sp}, in the Gaussian case for low-dimensional and high-dimensional synthesized data. We show that our modifications yield similar performance in low-dimensional settings and improved performance in high-dimensional settings.
     \item We release supplementary material including data and an R package that implements the proposed algorithm ... .
 \end{enumerate}
\section{Definitions and Concepts}\label{definitions}
Below we briefly list some of the most important concepts used in this paper.

	If there is an arrow from $a$ pointing towards $b$, $a$ is said to be a parent 
	of $b$. The set of parents of $b$ is denoted as $pa(b)$. If there is a bidirected edge between $a$ and $b$, $a$ and $b$ are said to be neighbors. The set of neighbors of a vertex $a$ is denoted as $ne(a)$. The expressions $pa(A)$ and $ne(A)$ denote the collection of  
	parents and neighbors of vertices in $A$ that are not themselves 
	elements of $A$. The boundary $bd(A)$ of a subset $A$ of vertices is the set of vertices in $V\setminus A$ that are parents or neighbors to vertices in $A$.

	A path of length $n$ from $a$ to $b$ is a sequence $a=a_0,\dots , a_n=b$ of 
	distinct vertices such that $(a_i\to a_{i+1})\in E$, for all $i=1,\dots ,n$. A chain of length $n$ from $a$ to $b$ is a sequence $a=a_0,\dots , a_n=b$ of 
	distinct vertices such that $(a_i\to a_{i+1})\in E$, or $(a_{i+1}\to a_i)\in E$, or $(a_{i+1}\leftrightarrow a_i)\in E$, for all $i=1,\dots ,n$. We say that $u$ is an ancestor of $v$ and $v$
	is a descendant of $u$ if there is a path from $u$ to $v$ in $G$.
	The set of ancestors of $v$ is denoted as $an(v)$, and we define $An(v) = an(v)\cup v$. We apply this definition to sets: $an(X) = \{\alpha | \alpha \textrm{ is an ancestor of } \beta \textrm{ for some } \beta \in X\}$.
	A partially directed cycle in a graph $G$ is a sequence of $n$ distinct vertices $v_1,\dots, v_n (n\ge 3)$,
	and $v_{n+1}\equiv v_1$, such that
 $\forall i (1\le i\le n)$ either $v_i\leftrightarrow v_{i+1}$ or $v_i\to v_{i+1}$, and
 $\exists j (1\le j\le n)$ such that $v_i\to v_{i+1}$.

	A graph with only undirected edges is called an undirected graph (UG). A graph with only
	directed edges and without directed cycles is called a directed acyclic graph (DAG). Acyclic directed mixed graphs, also known as semi-Markov(ian) \cite{pj}
	models contain directed ($\rightarrow$) and bidirected
	($\leftrightarrow$) edges subject to the restriction that there are no directed cycles \cite{r2}. A graph that has no partially directed cycles is called a \textit{chain graph}.

	A nonendpoint vertex $\zeta$ on a chain is a \emph{collider} on the chain if the edges preceding and succeeding $\zeta$ on the chain have an arrowhead at $\zeta$, that is, $\to \zeta \gets, or \leftrightarrow \zeta \leftrightarrow, or\leftrightarrow \zeta \gets, or\to \zeta \leftrightarrow$. A nonendpoint vertex $\zeta$ on a chain which is not a collider is a noncollider on the chain. A chain between vertices $\alpha$ and $\beta$ in  chain graph $G$ is said to be $m$-connecting given a set $Z$ (possibly empty), with $\alpha, \beta \notin Z$, if  every noncollider on the path is not in $Z$, and every collider on the path is in $An_G(Z)$.

	A chain that is not $m$-connecting given $Z$ is said to be blocked given (or by) $Z$.
	If there is no chain $m$-connecting $\alpha$ and $\beta$ given $Z$, then $\alpha$ and $\beta$ are said to be \emph{m-separated} given $Z$. Sets $X$ and $Y$ are $m$-separated given $Z$, if for every pair $\alpha, \beta$, with $\alpha\in X$ and $\beta \in Y$, $\alpha$ and $\beta$ are $m$-separated given $Z$ ($X$, $Y$, and $Z$ are disjoint sets; $X, Y$ are nonempty). We denote the independence model resulting from applying the $m$-separation criterion to $G$, by $\Im_m$(G). This is an extension of Pearl's $d$-separation criterion \cite{pearl1} to MVR chain graphs in that in a DAG $D$, a chain is $d$-connecting if and only if it is $m$-connecting.

We say that two MVR CGs $G$ and $H$
are Markov equivalent or that they are in the
same Markov equivalence class iff $\Im_m(G) = \Im_m(H)$.
If $G$ and $H$ have the same adjacencies and unshielded colliders, then $\Im_m(G) = \Im_m(H)$~\cite{ws}.

Just like for many other probabilistic graphical models there might exist multiple MVR CGs
that represent the same independence model. Sometimes it can however be desirable to have a unique graphical representation of the different representable independence models in the MVR CGs interpretation. 
A graph $G^*$ is said to be the essential MVR CG of an MVR CG $G$ if it has the same
skeleton as $G$ and contains all and only the arrowheads common to every MVR CG in the Markov equivalence class of $G$. One thing that can be noted here is that an essential MVR CG does not need to be a MVR CG. Instead these graphs can contain three
types of edges, undirected, directed and bidirected \cite{sonntagpena15}.

\section{Order-Dependent PC-like Algorithm}
In this section, we show that the PC-like algorithm proposed by Sonntag and Pe\~{n}a in \cite{sp} is order-dependent,
in the sense that the output can depend on the order in which the variables are given.
The PC-like algorithm for learning MVR CGs under the faithfulness assumption is formally described in Algorithm \ref{alg:MVRoriginalPC}.

\begin{algorithm}[!ht]
\caption{The order-dependent PC-like algorithm for learning MVR chain graphs \cite{sp}}\label{alg:MVRoriginalPC}
	\SetAlgoLined
	\small\KwIn{A set $V$ of nodes and a probability distribution $p$ faithful to an unknown MVR CG $G$ and an ordering order($V$) on the variables.}
	\KwOut{An MVR CG $G'$ s.t.
$G$ and $G'$ are Markov equivalent and $G'$ has exactly the minimum set of bidirected edges for its equivalence class.}
    Let $H$ denote the complete undirected graph over $V$\;
    \tcc{Skeleton Recovery}
\For{$i\gets 0$ \KwTo $|V_H|-2$}{
        \While{possible}{
            Select any ordered pair of nodes $u$ and $v$ in $H$ such that $u\in ad_H(v)$ and $|ad_H(u)\setminus v|\ge i$\, using order($V$);
            \tcc{$ad_H(x):=\{y\in V| x\erelbar{01} y, y\erelbar{01} x, \textrm{ or }x\erelbar{00}y\}$}
            \If{\textrm{there exists $S\subseteq (ad_H(u)\setminus v)$ s.t. $|S|=i$ and $u\perp\!\!\!\perp_p v|S$ (i.e., $u$ is independent of $v$ given $S$ in the probability distribution $p$)}}{
                Set $S_{uv} = S_{vu} = S$\;
                Remove the edge $u \erelbar{00} v$ from $H$\;
            }
        }
    }
    \tcc{$v$-structure Recovery}
    \For{\textrm{each $m$-separator $S_{uv}$}}{\If{\textrm{$u\erelbar{30} w\erelbar{03} v$ appears in the skeleton
	and $w$ is not in $S_{uv}$}}{
    \tcc{$u\erelbar{30} w$ means $u\erelbar{10} w$ or $u\erelbar{00} w$. Also, $w\erelbar{03} v$ means $w\erelbar{01} v$ or $w\erelbar{00}v.$}
    Determine a $v$-structure $u\erelbar{31} w\erelbar{13} v$\;
    }
    }
    Apply rules 1-3 in Figure \ref{Fig:rules} while possible\;
\tcc{After this line, the learned graph is the \textit{essential graph} of MVR CG $G$.}
Let $G'_u$ be the subgraph of $G'$ containing only the
nodes and the undirected edges in $G'$\;
Let $T$ be the junction tree of $G'_u$\; 
\tcc{If $G'_u$ is disconnected, the cliques belonging to different connected components can be linked with empty separators, as described in \cite[Theorem 4.8]{Golumbic}.}
Order the cliques $C_1,\cdots , C_n$ of $G'_u$ s.t. $C_1$ is the root of $T$ and if $C_i$ is closer to the root than $C_j$ in $T$ then $C_i < C_j$\;
Order the nodes such that if $A\in C_i$, $B\in C_j$, and $C_i < C_j$ then $A < B$\;
Orient the undirected edges in $G'$ according to the ordering obtained in line 21.
\end{algorithm}
\begin{figure}[!htbp]
	\centering
	\includegraphics[width=\linewidth]{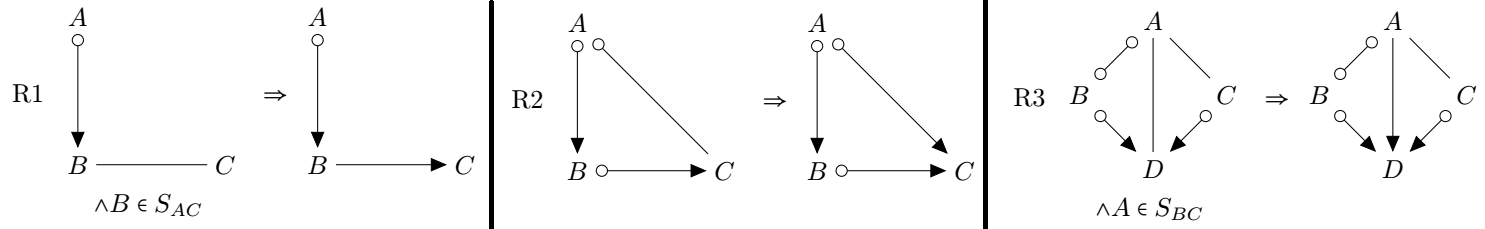}
	\caption{The Rules \cite{sp}} \label{Fig:rules}
\end{figure}
In applications, we do not have perfect conditional independence information.
Instead, we assume that we have an i.i.d. sample of size $n$ of variables $V = (X_1,\dots,Xp)$. In the PC-like algorithm \cite{sp} all conditional independence queries are estimated by statistical conditional independence tests at some pre-specified significance level (p value) $\alpha$. For example, if the distribution of $V$ is multivariate Gaussian, one can test for zero partial correlation, see, e.g., \cite{Kalisch07}. For this purpose, we use the $\mathsf{gaussCItest()}$ function from the R package \href{https://cran.r-project.org/web/packages/pcalg}{$\mathsf{pcalg}$} throughout this paper. Let order($V$) denote an ordering on the variables in $V$. We now consider the role of
order($V$) in every step of the algorithm.

In the skeleton recovery phase of the PC-like algorithm \cite{sp}, the order of variables affects the estimation of the skeleton and the separating sets. In particular, as noted for the special case of Bayesian networks in~\cite{Colombo2014}, for each level of $i$, the order of variables determines the order in which pairs of adjacent
vertices and subsets $S$ of their adjacency sets are considered (see lines 4 and 5 in Algorithm \ref{alg:MVRoriginalPC}). The skeleton $H$ is updated after each edge removal. Hence, the adjacency sets typically change within one level of $i$, and this affects which other conditional independencies are
checked, since the algorithm only conditions on subsets of the adjacency sets. When we have perfect conditional independence information,  all orderings on the variables lead to the same output. In the sample version, however, we typically make
mistakes in keeping or removing edges, because conditional independence relationships have to be estimated from data. In such cases, the resulting changes in the adjacency
sets can lead to different skeletons, as illustrated in Example \ref{ex1OrderDepMVR}.

Moreover, different variable orderings can lead to different separating sets in the skeleton recovery phase.
When we have perfect conditional independence information, this is not important, because any valid separating set leads to the
correct \textit{v}-structure decision in the orientation phase. In the sample version, however, different separating
sets in the skeleton recovery phase of the algorithm may yield different decisions about \textit{v}-structures in the orientation phase.
This is illustrated in Example \ref{ex2OrderDepMVR}.

Finally, we consider the role of order($V$) on the orientation rules in the essential graph recovery phase
of the sample version of the PC-like algorithm. Example \ref{ex:orientdepMVR} illustrates that different variable orderings can lead to different orientations, even if the skeleton and separating sets are order-independent.

\begin{example}[Order-dependent skeleton of the PC-like algorithm.]\label{ex1OrderDepMVR}
Suppose that the distribution of $V = \{a,b,c,d,e\}$ is faithful to the DAG in Figure
\ref{fig:OrderDepex1MVR}(a). This DAG encodes the following conditional independencies (using the notation defined in line 5 of Algorithm \ref{alg:MVRoriginalPC}) with minimal separating
sets: $a\perp\!\!\!\perp d|\{b,c\}$ and $a\perp\!\!\!\perp e|\{b,c\}$.

Suppose that we have an i.i.d. sample of $(a,b,c,d,e)$, and that the following
conditional independencies with minimal separating sets are judged to hold at some significance level $\alpha$: $a\perp\!\!\!\perp d|\{b,c\}$, $a\perp\!\!\!\perp e|\{b,c,d\}$, and $c\perp\!\!\!\perp e|\{a,b,d\}$. Thus, the first two are correct, while the third is false.

We now apply the skeleton recovery phase of the PC-like algorithm with two different orderings: $\textrm{order}_1(V)=(d,e,a,c,b)$ and $\textrm{order}_2(V)=(d,c,e,a,b)$. The resulting skeletons are shown in Figures \ref{fig:OrderDepex1MVR}(b) and \ref{fig:OrderDepex1MVR}(c), respectively. 
\begin{figure}[!htpb]
    \centering
	\[\resizebox{\textwidth}{!}{\begin{tikzpicture}[transform shape]
	\tikzset{vertex/.style = {shape=circle,draw,minimum size=1em}}
	\tikzset{edge/.style = {->,> = latex'}}
	\node[vertex] (o) at  (0,.5) {$e$};
	\node[vertex] (p) at  (0,2) {$d$};
	\node[vertex] (q) at  (0,4) {$a$};
	\node[vertex] (r) at  (-2,3) {$b$};
	\node[vertex] (s) at  (2,3) {$c$};
	\node (t) at (0,-.5) {$(a)$};
	\draw[edge] (q) to (r);
	\draw[edge] (q) to (s);
	\draw[edge] (r) to (s);
	\draw[edge] (r) to (p);
	\draw[edge] (r) to (o);
	\draw[edge] (s) to (p);
	\draw[edge] (s) to (o);
	\draw[edge] (p) to (o);
	
	\node[vertex] (i) at  (5,0.5) {$e$};
	\node[vertex] (j) at  (5,2) {$d$};
	\node[vertex] (k) at  (5,4) {$a$};
	\node[vertex] (l) at  (3,3) {$b$};
	\node[vertex] (m) at  (7,3) {$c$};
	\node (n) at (5,-.5) {$(b)$};
	\draw (j) to (l);
	\draw (j) to (i);
	\draw (i) to (l);
	\draw (k) to (m);
	\draw (k) to (l);
	\draw (j) to (m);
	\draw (l) to (m);
	
	\node[vertex] (e) at  (10,0.5) {$e$};
	\node[vertex] (d) at  (10,2) {$d$};
	\node[vertex] (a) at  (10,4) {$a$};
	\node[vertex] (b) at  (8,3) {$b$};
	\node[vertex] (c) at  (12,3) {$c$};
	\node (f) at (10,-.5) {$(c)$};
	\draw (c) to (d);
	\draw (e) to (d);
	\draw (e) to (b);
	\draw (a) to [out=300,in=60] (e);
	\draw (b) to (c);
	\draw (a) to (b);
	\draw (a) to (c);
	\draw (b) to (d);
	\end{tikzpicture}}\]
    \caption{(a) The DAG $G$, (b) the skeleton returned by  Algorithm \ref{alg:MVRoriginalPC} with $\textrm{order}_1(V)$, (c) the skeleton returned by  Algorithm \ref{alg:MVRoriginalPC} with $\textrm{order}_2(V)$. 
}
    \label{fig:OrderDepex1MVR}
\end{figure}
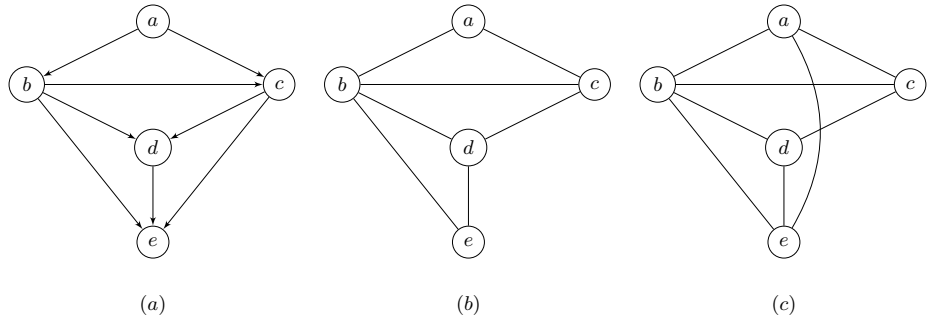

We see that the skeletons are different, and that both are incorrect as the edge $c\erelbar{00}e$ is missing. The skeleton for $\textrm{order}_2(V)$ contains an additional
error, as there is an additional edge $a\erelbar{00}e$. We now go through Algorithm \ref{alg:MVRoriginalPC} to see what happened. We start with a complete undirected graph on $V$. When $i= 0$, variables are tested for marginal independence, and the algorithm correctly does not remove any edge. Also, when $i=1$, the algorithm correctly does not remove any edge. When $i= 2$, there is a pair of vertices that is thought to be conditionally independent given a subset of size two, and the
algorithm correctly removes the edge between $a$ and $d$. When $i=3$, there are two pairs of vertices that are thought to be conditionally independent given a subset of size three. Table \ref{t1OrderDepMVRex1}
shows the trace table of Algorithm \ref{alg:MVRoriginalPC} for $i=3$ and $\textrm{order}_1(V)=(d,e,a,c,b)$.
\begin{table}[!htpb]
\caption{The trace table of Algorithm \ref{alg:MVRoriginalPC} for $i=3$ and $\textrm{order}_1(V)=(d,e,a,c,b)$.}\label{t1OrderDepMVRex1}
\centering
\begin{tabular}{c|c|c|c|c}
 Ordered Pair $(u,v)$& $ad_H(u)$ & $S_{uv}$&Is $S_{uv}\subseteq ad_H(u)\setminus\{v\}$?& Is $u\erelbar{00} v$ removed?\\
\midrule
\midrule
   $(e,a)$ & $\{a,b,c,d\}$&$\{b,c,d\}$&	Yes&	Yes\\
    \midrule
  $(e,c)$  &$\{b,c,d\}$&$\{a,b,d\}$&	No&	No \\
\midrule
$(c,e)$ &$\{a,b,d,e\}$ &$\{a,b,d\}$&Yes&	Yes\\
\bottomrule
\end{tabular}
\end{table}

Table \ref{t2OrderDepMVRex1}
shows the trace table of Algorithm \ref{alg:MVRoriginalPC} for $i=3$ and $\textrm{order}_2(V)=(d,c,e,a,b)$.

\begin{table}[!htpb]
\caption{The trace table of Algorithm \ref{alg:MVRoriginalPC} for $i=3$ and $\textrm{order}_2(V)=(d,c,e,a,b)$.}
\centering
\begin{tabular}{c|c|c|c|c}
 Ordered Pair $(u,v)$& $ad_H(u)$ & $S_{uv}$&Is $S_{uv}\subseteq ad_H(u)\setminus\{v\}$?& Is $u\erelbar{00} v$ removed?\\
\midrule
\midrule
   $(c,e)$ &$\{a,b,d,e\}$ &$\{a,b,d\}$&Yes&	Yes\\ 
    \midrule
  $(e,a)$ & $\{a,b,d\}$&$\{b,c,d\}$&	No&	No\\
    \midrule
  $(a,e)$  &$\{b,c,e\}$&$\{b,c,d\}$&No&No\\
\bottomrule
\end{tabular}\label{t2OrderDepMVRex1}
\end{table}
\end{example}

\begin{example}[Order-dependent separating sets and \textit{v}-structures of the PC-like algorithm.]\label{ex2OrderDepMVR}
Suppose that the distribution of $V = \{a,b,c,d,e\}$ is faithful to the DAG in Figure
\ref{fig:OrderDepex2MVR}(a). This DAG encodes the following conditional independencies with minimal separating sets: $a\perp\!\!\!\perp d|b, a\perp\!\!\!\perp e|\{b,c\}, a\perp\!\!\!\perp e|\{c,d\}, b\perp\!\!\!\perp c, b\perp\!\!\!\perp e|d,$ and $c\perp\!\!\!\perp d$.

Suppose that we have an i.i.d. sample of $(a,b,c,d,e)$. Assume that all true conditional independencies are judged to hold except $c\perp\!\!\!\perp d$. Suppose that $c\perp\!\!\!\perp d|b$ and $c\perp\!\!\!\perp d|e$ are thought to hold. Thus, the first is correct, while the second is false. We now apply the \textit{v}-structure recovery phase of the PC-like algorithm with two different orderings: $\textrm{order}_1(V)=(d,c,b,a,e)$ and $\textrm{order}_3(V)=(c,d,e,a,b)$. The resulting CGs are shown in Figures \ref{fig:OrderDepex2MVR}(b) and \ref{fig:OrderDepex2MVR}(c), respectively. Note that while the separating set for vertices $c$ and $d$ with $\textrm{order}_1(V)$ is $S_{dc}=S_{cd}=\{b\}$, the separating set for them with $\textrm{order}_2(V)$ is $S_{cd}=S_{dc}=\{e\}$.
\begin{figure}[!htbp]
    \centering
	\[\begin{tikzpicture}[transform shape]
	\tikzset{vertex/.style = {shape=circle,draw,minimum size=1em}}
	\tikzset{edge/.style = {->,> = latex'}}
	\node[vertex] (o) at  (1,.5) {$e$};
	\node[vertex] (p) at  (-1,.5) {$d$};
	\node[vertex] (q) at  (0,2) {$a$};
	\node[vertex] (r) at  (-1,1.5) {$b$};
	\node[vertex] (s) at  (1,1.5) {$c$};
	\node (t) at (0,0) {$(a)$};
	\draw[edge] (r) to (q);
	\draw[edge] (s) to (q);
	\draw[edge] (r) to (p);
	\draw[edge] (s) to (o);
	\draw[edge] (p) to (o);
	
	\node[vertex] (i) at  (5,.5) {$e$};
	\node[vertex] (j) at  (3,.5) {$d$};
	\node[vertex] (k) at  (4,2) {$a$};
	\node[vertex] (l) at  (3,1.5) {$b$};
	\node[vertex] (m) at  (5,1.5) {$c$};
	\node (n) at (4,0) {$(b)$};
	\draw[edge] (l) to (k);
	\draw[edge] (j) to (i);
	\draw[edge] (m) to (k);
	\draw[edge] (m) to (i);
	\draw (j) to (l);
	
	\node[vertex] (e) at  (9,.5) {$e$};
	\node[vertex] (d) at  (7,.5) {$d$};
	\node[vertex] (a) at  (8,2) {$a$};
	\node[vertex] (b) at  (7,1.5) {$b$};
	\node[vertex] (c) at  (9,1.5) {$c$};
	\node (f) at (8,0) {$(c)$};
	\draw[edge] (b) to (a);
	\draw[edge] (c) to (a);
	\draw (e) to (d);
	\draw (e) to (c);
	\draw (b) to (d);
	\end{tikzpicture}\]
    \caption{(a) The DAG $G$, (b) the CG returned after the \textit{v}-structure recovery phase of  Algorithm \ref{alg:MVRoriginalPC} with $\textrm{order}_1(V)$, (c) the CG returned after the \textit{v}-structure recovery phase of  Algorithm \ref{alg:MVRoriginalPC} with $\textrm{order}_3(V)$.}\label{fig:OrderDepex2MVR}
\end{figure}
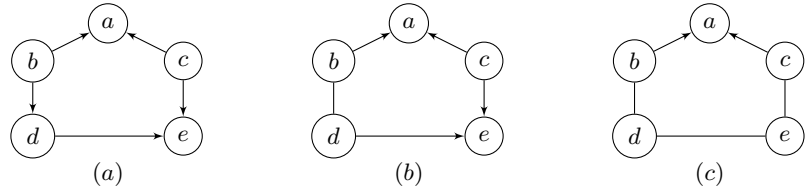

This illustrates that order-dependent separating sets in the skeleton recovery phase of the sample version of the PC-algorithm can lead to order-dependent \textit{v}-structures.
\end{example}
\begin{example}[Order-dependent orientation rules of the PC-like algorithm.]\label{ex:orientdepMVR}
Consider the graph in Figure \ref{fig:orientdepMVR}, and assume that this is the output of the sample version of the PC-like algorithm after  \textit{v}-structure recovery. Also, consider that $c\in S_{a,d}$ and $d\in S_{b,f}$. Thus, we have two $v$-structures, namely $a\erelbar{01}c\erelbar{10}e$ and $b\erelbar{01}d\erelbar{10}f$, and four unshielded triples, namely $(e,c,d), (c,d,f), (a,c,d),$ and $(b,d,c)$. Thus, we then apply the orientation rules
in the essential recovery phase of the algorithm, starting with rule R1. If one of the two unshielded triples $(e,c,d)$ or $(a,c,d)$ is considered first, we obtain $c\erelbar{01}d$. On the other hand, if
one of the unshielded triples $(b,d,c)$ or $(c,d,f)$ is considered first, then we obtain
$c\erelbar{10}d$. Note that we have no issues with overwriting of edges here, since as soon as the
edge $c\erelbar{00}d$ is oriented, all edges are oriented and no further orientation rules are applied.
These examples illustrate that the essential graph recovery phase of the PC-like algorithm can be order-dependent
regardless of the output of the previous steps.
\begin{figure}[!ht]
    \centering
	\[\begin{tikzpicture}[transform shape]
	\tikzset{vertex/.style = {shape=circle,draw,minimum size=1em}}
	\tikzset{edge/.style = {->,> = latex'}}
	\node[vertex] (e) at  (-1.5,0) {$e$};
	\node[vertex] (d) at  (2,0) {$d$};
	\node[vertex] (a) at  (0,1) {$a$};
	\node[vertex] (b) at  (2,1) {$b$};
	\node[vertex] (c) at  (0,0) {$c$};
	\node[vertex] (f) at  (3.5,0) {$f$};
	\draw[edge] (a) to (c);
	\draw[edge] (b) to (d);
	\draw[edge] (e) to (c);
	\draw[edge] (f) to (d);
	\draw (c) to (d);
	\end{tikzpicture}\]
    \caption{Possible mixed graph after \textit{v}-structure recovery phase of the sample version of the PC-like algorithm.}\label{fig:orientdepMVR}
\end{figure}
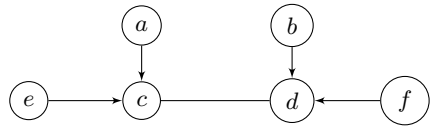
\end{example}

\section{Order Independent Algorithms for Learning MVR CGs}
We now propose several modifications of the original PC-like algorithm (and hence also of the related algorithms) that remove the order-dependence in the various stages of the algorithm, analogously to what Colombo and Maathuis~\cite{Colombo2014} did for the original PC algorithm in the case of DAGs. For this purpose, we discuss the skeleton, \textit{v}-structures, and the orientation rules, respectively.
\subsection{Order-Independent Skeleton Recovery}
We first consider estimation of the skeleton in the adjacency search of the PC-like algorithm. The pseudocode for our modification is given in Algorithm \ref{alg:MVRstablePC}. The resulting PC-like algorithm in Algorithm \ref{alg:MVRstablePC} is called \textit{stable PC-like}.

The main difference between Algorithms \ref{alg:MVRoriginalPC} and \ref{alg:MVRstablePC} is given by the for-loop on lines
3-5 in the latter one, which computes and stores the adjacency sets $a_H(v_i)$ of all variables
after each new size $i$ of the conditioning sets. These stored adjacency sets $a_H(v_i)$ are used
whenever we search for conditioning sets of this given size $i$. Consequently, an edge deletion
on line 10 no longer affects which conditional independencies are checked for other pairs of
variables at this level of $i$.

In other words, at each level of $i$, Algorithm \ref{alg:MVRstablePC} records which edges should be removed, but for the purpose of the adjacency sets it removes these edges only when it goes to the
next value of $i$. Besides resolving the order-dependence in the estimation of the skeleton,
our algorithm has the advantage that it is easily parallelizable at each level of $i$.
The stable PC-like algorithm is 
correct, i.e. it returns an MVR CG to which the given probability distribution is faithful (Theorem \ref{thm:correctPCstableMVR}), and it yields order-independent skeletons in the sample version (Theorem \ref{thm:stableskeletonMVR}). We illustrate the algorithm in Example \ref{ex:stableskeletonsMVR}.
\begin{algorithm}[!ht]
\caption{The order-independent (stable) PC-like algorithm for learning MVR chain graphs.}\label{alg:MVRstablePC}
	\SetAlgoLined
	\small\KwIn{A set $V$ of nodes and a probability distribution $p$ faithful to an unknown MVR CG $G$ and an ordering order($V$) on the variables.}
	\KwOut{An MVR CG $G'$ s.t.
$G$ and $G'$ are Markov equivalent and $G'$ has exactly the minimum set of bidirected edges for its equivalence class.}
    Let $H$ denote the complete undirected graph over $V=\{v_1,\dots,v_n\}$\;
\tcc{Skeleton Recovery}
\For{$i\gets 0$ \KwTo $|V_H|-2$}{
    \For{$j\gets 1$ \KwTo $|V_H|$}{
        Set $a_H(v_i)=ad_H(v_i)$\;
    }
        \While{possible}{
            Select any ordered pair of nodes $u$ and $v$ in $H$ such that $u\in a_H(v)$ and $|a_H(u)\setminus v|\ge i$\, using order($V$);
            
            \If{\textrm{there exists $S\subseteq (a_H(u)\setminus v)$ s.t. $|S|=i$ and $u\perp\!\!\!\perp_p v|S$ (i.e., $u$ is independent of $v$ given $S$ in the probability distribution $p$)}}{
                Set $S_{uv} = S_{vu} = S$\;
                Remove the edge $u \erelbar{00} v$ from $H$\;
            }
        }
    }
    \tcc{$v$-structure Recovery and orientation rules}
    Follow the same procedures in Algorithm \ref{alg:MVRoriginalPC} (lines: 11-21). 
\end{algorithm}
\begin{theorem}\label{thm:correctPCstableMVR}
    Let the distribution of $V$ be faithful to an MVR CG $G$, and assume that we are given perfect conditional independence information about all pairs of variables $(u,v)$ in $V$ given subsets $S\subseteq V\setminus \{u,v\}$. Then the output of the stable PC-like algorithm is an MVR CG that has exactly the minimum set of bidirected edges for its equivalence class.
\end{theorem}

\begin{theorem}\label{thm:stableskeletonMVR}
    The skeleton resulting from the sample version of the stable PC-like algorithm is order-independent.
\end{theorem}


\begin{example}[Order-independent skeletons]\label{ex:stableskeletonsMVR}
We go back to Example \ref{ex1OrderDepMVR}, and consider the sample version of Algorithm \ref{alg:MVRstablePC}. The algorithm now outputs the skeleton shown in Figure \ref{fig:OrderDepex1MVR}(b) for both orderings $\textrm{order}_1(V)$ and $\textrm{order}_2(V)$.

We again go through the algorithm step by step. We start with a complete undirected
graph on $V$. No conditional independence found when $i=0$. Also, when $i=1$, the algorithm correctly does not remove any edge. When $i= 2$, the algorithm first computes the new adjacency sets: $a_H(v)=V\setminus\{v\}, \forall v\in V$. There is a pair of variables that is thought to be conditionally independent given a subset of size two, namely $(b,c)$. Since the sets $a_H(v)$ are not updated after
edge removals, it does not matter in which order we consider the ordered pair. Any ordering leads to the removal of edge between $b$ and $c$.  When $i= 3$, the algorithm first computes the new adjacency sets: $a_H(b)=a_H(c)=\{a,d,e\}$ and $a_H(v)=V\setminus\{v\}, \textrm{ for } v=a,d,e$. There are two pairs of variables that are thought to be conditionally independent given a subset of size three, namely $(a,e)$ and $(c,e)$. Since the sets $a_H(v)$ are not updated after
edge removals, it does not matter in which order we consider the ordered pair. Any ordering leads to the removal of both edges $a\erelbar{00}e$ and $c\erelbar{00}e$. 
\end{example}

\subsection{Order-Independent \textit{v}-structures Recovery}
We propose two methods to resolve the order-dependence in the determination of the \textit{v}-structures, using the conservative PC algorithm (CPC) of Ramsey et al. \cite{Ramsey:2006} and the majority rule PC-like algorithm (MPC) of Colombo \& Maathuis~\cite{Colombo2014}.

The \textbf{Conservative PC-like algorithm  (CPC-like algorithm)} works as follows. Let $H$ be the undirected graph resulting from the skeleton recovery phase  of the PC-like algorithm (Algorithm \ref{alg:MVRoriginalPC}). For all unshielded triples $(X_i, X_j, X_k)$ in $H$, determine all subsets
$S$ of $ad_H(X_i)$ and of $ad_H(X_k)$ that make $X_i$ and $X_k$ conditionally independent, i.e., that
satisfy $X_i\perp\!\!\!\perp_p X_k|S$. We refer to such sets as separating sets. The triple $(X_i, X_j, X_k)$ is labelled as \textit{unambiguous} if at least one such separating set is found and either $X_j$ is in all separating sets or in none of them; otherwise it is labelled as \textit{ambiguous}. If the triple is
unambiguous, it is oriented as \textit{v}-structure if and only if $X_j$ is in none of the separating
sets. Moreover, in the \textit{v}-structure recovery phase of the PC-like algorithm (Algorithm \ref{alg:MVRoriginalPC}, lines 11-15), the orientation rules are
adapted so that only unambiguous triples are oriented. The output of the CPC-like algorithm
is a mixed graph in which ambiguous triples are marked. 
We refer to the combination of the stable PC-like and CPC-like algorithms as the \textit{stable CPC-like algorithm}.

In the case of DAGs, Colombo and Maathuis \cite{Colombo2014} found that the CPC-algorithm can be very conservative, in the sense that very few
unshielded triples are unambiguous in the sample version,  where conditional independence relationships have to be estimated from
data. They proposed a minor
modification of the CPC approach, called \textit{Majority rule PC algorithm (MPC)} to mitigate the (unnecessary) severity of CPC-like approach.  We similarly propose the \textbf{Majority rule PC-like algorithm (MPC-like)} for MVR CGs. As in the CPC-like algorithm, we
first determine all subsets $S$ of $ad_H(X_i)$ and of $ad_H(X_k)$ that make $X_i$ and $X_k$ conditionally independent, i.e., that
satisfy $X_i\perp\!\!\!\perp_p X_k|S$. The triple $(X_i, X_j, X_k)$ is labelled as \textit{($\alpha, \beta$)-unambiguous} if at least one such separating set is found
or $X_j$ is in no more than $\alpha$\% or no less than $\beta$\% of the separating sets, for $0 \le \alpha\le \beta \le 100$. Otherwise it is labelled as \textit{ambiguous}. (As an example, consider $\alpha = 30$ and $\beta = 60$.) If a triple is unambiguous, it is oriented as a \textit{v}-structure if and only if $X_j$ is in less than $\alpha$\% of the separating sets. As in the CPC-like algorithm, the orientation rules in the \textit{v}-structure recovery phase of the PC-like algorithm (Algorithm \ref{alg:MVRoriginalPC}, lines 11-15) are adapted so
that only unambiguous triples are oriented, and the output is a mixed graph in
which ambiguous triples are marked.
Note that the CPC-like algorithm is the special case of the MPC-like algorithm with $\alpha = 0$ and $\beta = 100$.
We refer to the combination of the stable PC-like and MPC-like algorithms as the \textit{stable MPC-like algorithm}.

\begin{theorem}\label{thm:correctCPCstableMVR}
    Let the distribution of $V$ be faithful to an MVR CG $G$, and assume that we are given perfect conditional independence information about all pairs of variables $(u,v)$ in $V$ given subsets $S\subseteq V\setminus \{u,v\}$. Then the output of the (stable) CPC/MPC-like algorithm is an MVR CG that is Markov equivalent with $G$ that has exactly the minimum set of bidirected edges for its equivalence class.
\end{theorem}
\begin{theorem}\label{thm:stablevstructs}
    The decisions about \textit{v}-structures in the sample version of the stable CPC/MPC-like algorithm is order-independent.
\end{theorem}
\begin{example}[Order-independent decisions about \textit{v}-structures]\label{ex:stablevstruct}
We consider the sample versions of the stable CPC/MPC-like algorithm, using the same input as in Example \ref{ex2OrderDepMVR}. In particular, we assume that all conditional independencies induced by the MVR CG in Figure \ref{fig:OrderDepex2MVR}(a)
are judged to hold except $c\perp\!\!\!\perp d$. Suppose that $c\perp\!\!\!\perp d|b$ and $c\perp\!\!\!\perp d|e$ are thought to hold. Let $\alpha=\beta=50$. 

Denote the skeleton after the skeleton recovery phase by $H$. We consider the unshielded triple $(c,e,d)$.
First, we compute $a_H(c)=\{a,d,e\}$ and $a_H(d)=\{a,b,c,e\}$, when $i=1$. We now consider all
subsets $S$ of these adjacency sets, and check whether $c\perp\!\!\!\perp d|S$. The following separating sets are found: $\{b\},\{e\}$, and $\{b,e\}$. Since $e$ is in some but not all of these separating sets, the stable CPC-like algorithm determines that the triple is ambiguous, and no orientations are performed. Since $e$ is in more than half of the
separating sets, stable MPC-like determines that the triple is unambiguous and not a \textit{v}-structure. The output of both algorithms is given in Figure \ref{fig:OrderDepex2MVR}(c).
\end{example}
At this point it should be clear why the modified PC-like algorithm is labeled ``conservative": it is more cautious than the (stable) PC-like algorithm in drawing unambiguous conclusions about orientations. As we showed in Example \ref{ex:stablevstruct}, the output of the (stable) CPC-like algorithm may
not be collider equivalent with the true MVR CG $G$, if the resulting CG contains an ambiguous triple. 
\subsection{Order-Independent Orientation Rules}
Even when the skeleton and the determination of the \textit{v}-structures are order-independent,
Example \ref{ex:orientdepMVR} showed that there might be some order-dependent steps left in the sample version. Regarding the orientation rules, we note that
the PC-like algorithm does not suffer from conflicting \textit{v}-structures (as shown in \cite{Colombo2014}
for the PC-algorithm in the case of DAGs), because bi-directed edges are allowed. However, the three orientation rules still suffer from order-dependence issues (see Example \ref{ex:orientdepMVR} and Figure \ref{fig:orientdepMVR}). To solve this problem, we can  use lists of candidate edges for each orientation rule as follows: we first generate a list of all edges that can be oriented by rule R1. We orient all these edges, creating bi-directed edges if there are conflicts. We do the same for rules
R2 and R3, and iterate this procedure until no more edges can be oriented.

When using this procedure, we add the letter $L$ (standing for lists), e.g., (stable) LCPC-like
and (stable) LMPC-like. The (stable) LCPC-like
and (stable) LMPC-like algorithms are fully order-independent in the sample versions. The
procedure is illustrated in Example \ref{ex:orientstableMVR}.
\begin{theorem}\label{thm:correctLCPCstableMVR}
    Let the distribution of $V$ be faithful to an MVR CG $G$, and assume that we are given perfect conditional independence information about all pairs of variables $(u,v)$ in $V$ given subsets $S\subseteq V\setminus \{u,v\}$. Then the output of the (stable) LCPC/LMPC-like algorithm is an MVR CG that is Markov equivalent with $G$ that has exactly the minimum set of bidirected edges for its equivalence class.
\end{theorem}
\begin{theorem}\label{thm:correctLCPCMVR}
    The sample versions of (stable) CPC-like
and (stable) MPC-like algorithms are fully order-independent.
\end{theorem}

 Table \ref{Relations:modifiedalgs} summarizes the three order-dependence issues explained above and the corresponding modifications of the PC-like algorithm that removes the given order-dependence problem.
\begin{table}[!ht]
\caption{Order-dependence issues and corresponding modifications of the PC-like algorithm
that remove the problem. ``Yes" indicates that the corresponding aspect
of the graph is estimated order-independently in the sample version.}\label{Relations:modifiedalgs}
\centering
\resizebox{!}{.07\textheight}{
\begin{tabular}{c|c|c|c|}
 & skeleton & \textit{v}-structures decisions & edges orientations\\
\midrule
\midrule
PC-like & No & No & No\\
    \midrule
stable PC-like & Yes & No & No \\
\midrule
stable CPC/MPC-like & Yes & Yes & No \\
\midrule
stable LCPC/LMPC-like & Yes & Yes & Yes\\
\bottomrule
\end{tabular}}
\end{table}
\begin{example}\label{ex:orientstableMVR}
Consider the structure shown in Figure \ref{fig:orientdepMVR}. As a first step, we construct a list containing all candidate structures eligible for orientation rule R1 in the phase of the essential graph recovery. The list
contains the unshielded triples $(e,c,d), (c,d,f), (a,c,d),$ and $(b,d,c)$.
Now, we go through each element in the list and we orient the edges accordingly, allowing
bi-directed edges. This yields the edge orientation $c\erelbar{11} d$, regardless of the ordering of
the variables.
\end{example}
\section{Evaluation}
In this section, we compare the performance of our algorithms (Table \ref{Relations:modifiedalgs}) with the original PC-like
learning algorithm by running them
on randomly generated MVR chain graphs in low-dimensional and high-dimensional data, respectively.  We report on the Gaussian case only because of space limitations.

\subsubsection{Performance Evaluation on Random MVR CGs (Gaussian case)}

To investigate the performance of the proposed learning methods in this paper, we use the same approach that \cite{mxg} used in evaluating the performance of the LCD algorithm on LWF chain graphs. We run our algorithms on randomly generated MVR chain graphs and then we compare the results and report summary error measures in all cases.

\subsubsection{Data Generation Procedure}
First we explain the way in which the random MVR chain graphs and random samples are generated.
Given a vertex set $V$, let $p = |V|$ and $N$ denote the average degree of edges (including bidirected, pointing out, and pointing in) for each vertex. We generate a random MVR chain graph on $V$ as
follows:
\begin{itemize}
	\item Order the $p$ vertices and initialize a $p\times p$ adjacency matrix $A$ with zeros;
	\item Set each element in the lower triangle part of $A$ to be a random number generated from a Bernoulli distribution with probability of occurrence $s = N/(p-1)$;
	\item Symmetrize $A$ according to its lower triangle;
	\item Select an integer $k$ randomly from $\{1,\dots,p\}$ as the number of chain components;
	
	\item Split the interval $[1, p]$ into $k$ equal-length subintervals $I_1,\dots,I_k$ so that the set of variables into each subinterval $I_m$ forms a chain component $C_m$; 
	
	\item Set $A_{ij} = 0$ for any $(i, j)$ pair such that $i \in I_l, j \in I_m$ with $l > m$.
\end{itemize}

This procedure yields an adjacency matrix $A$ for a chain graph with $(A_{ij} = A_{ji} = 1)$ representing a bidirected edge between $V_i$ and $V_j$ and $(A_{ij} =1,  A_{ji} =0)$ representing a directed edge
from $V_i$ to $V_j$. Moreover, it is not difficult to see that $\mathbb{E}[\textrm{vertex degree}] = N$, where an adjacent vertex can
be linked by either a bidirected or a directed edge.
In order to sample the artificial CGs, we first transform
them into DAGs and then generate samples from these DAGs under marginalization, as indicated in \cite{jv1}, using \href{https://www.hugin.com/}{Hugin}.
\subsubsection{Experimental Results}
We evaluate the performance of the proposed algorithms in terms of the six measurements that are commonly used for constraint-based learning algorithms: (a) the true positive
rate (TPR) (also known as sensitivity, recall, and hit rate), (b) the false positive rate (FPR) (also known as fall-out), (c) the true discovery rate (TDR) (also known as precision or positive predictive value), (d) accuracy (ACC) for the skeleton, (e) the structural Hamming distance (SHD) (this is the metric described in \cite{Tsamardinos2006} to  compare the
structure of the learned and the original graphs), and (f) run-time for the LCG recovery algorithms. 
 In short, $TPR=\frac{\textrm{true positive } (TP)}{\textrm{the number of positive cases in the data } (Pos)}$ is the ratio of  the number of correctly identified edges over total number of edges  (in true graph), $FPR=\frac{\textrm{false positive }(FP)}{\textrm{the number of negative cases in the data }(Neg)}$ is the ratio of the number of incorrectly identified edges over total number of gaps, $TDR=\frac{\textrm{true positive } (TP)}{\textrm{the total number of edges in the recovered CG}}$ is the ratio of  the number of correctly identified edges over total number of edges (both in estimated graph), $ACC=\frac{\textrm{true positive }(TP) +\textrm{ true negative }(TN)}{Pos+Neg}$ and
 $SHD$ is the number of legitimate operations needed to change the current resulting graph to the true essential graph,
 where legitimate operations are: (a) add or delete an edge and (b) insert, delete or reverse an edge
 orientation. 
In principle, large values of TPR, TDR, and ACC, and small values of FPR and SHD indicate good performance. All of these six measurements are computed on the essential graphs of the CGs, rather than the CGs directly, to avoid spurious differences due to random orientation of undirected edges.

 In our simulation, for low-dimensional settings, we set $N$ (expected number of adjacent vertices) to 2 and change the parameters $p$ (the number of vertices) and $n$ (sample size) and
  as follows:
 \begin{itemize}
     \item $p\in\{10, 20, 30, 40, 50\}$,
     \item $n\in\{500, 1000, 5000, 10000\}$.
 \end{itemize}

 For each $(p,N)$ combination, we first generate 30 random MVR CGs. We then generate a random Gaussian distribution based on each graph (transformed DAG) and draw an identically independently distributed
 (i.i.d.) sample of size $n$ from this distribution for each possible $n$. For each sample, four different
 significance levels $(\alpha = 0.001, 0.005, 0,01, 0.05)$ are used to perform the hypothesis tests. The \textit{null hypothesis} $H_0$ is ``two variables $u$ and $v$ are conditionally independent given a set $C$ of variables" and alternative $H_1$ is that $H_0$ may not hold. We then compare the results to access the influence of the significance testing level on the performance of our algorithms.

 For the high-dimensional setting, we generate 30 random MVR CGs with 1000 vertices for which the expected number of adjacent vertices for each vertex is 2. We then generate a random Gaussian distribution based on each graph (transformed DAG) and draw an identically independently distributed
 (i.i.d.) sample of size 50 from this distribution for each DAG. These numbers are similar to ones that could be encountered in gene regulatory network experiments~\cite[section 6]{Colombo2014}.

\begin{figure}
	\centering
	\includegraphics[scale=.22,page=1]{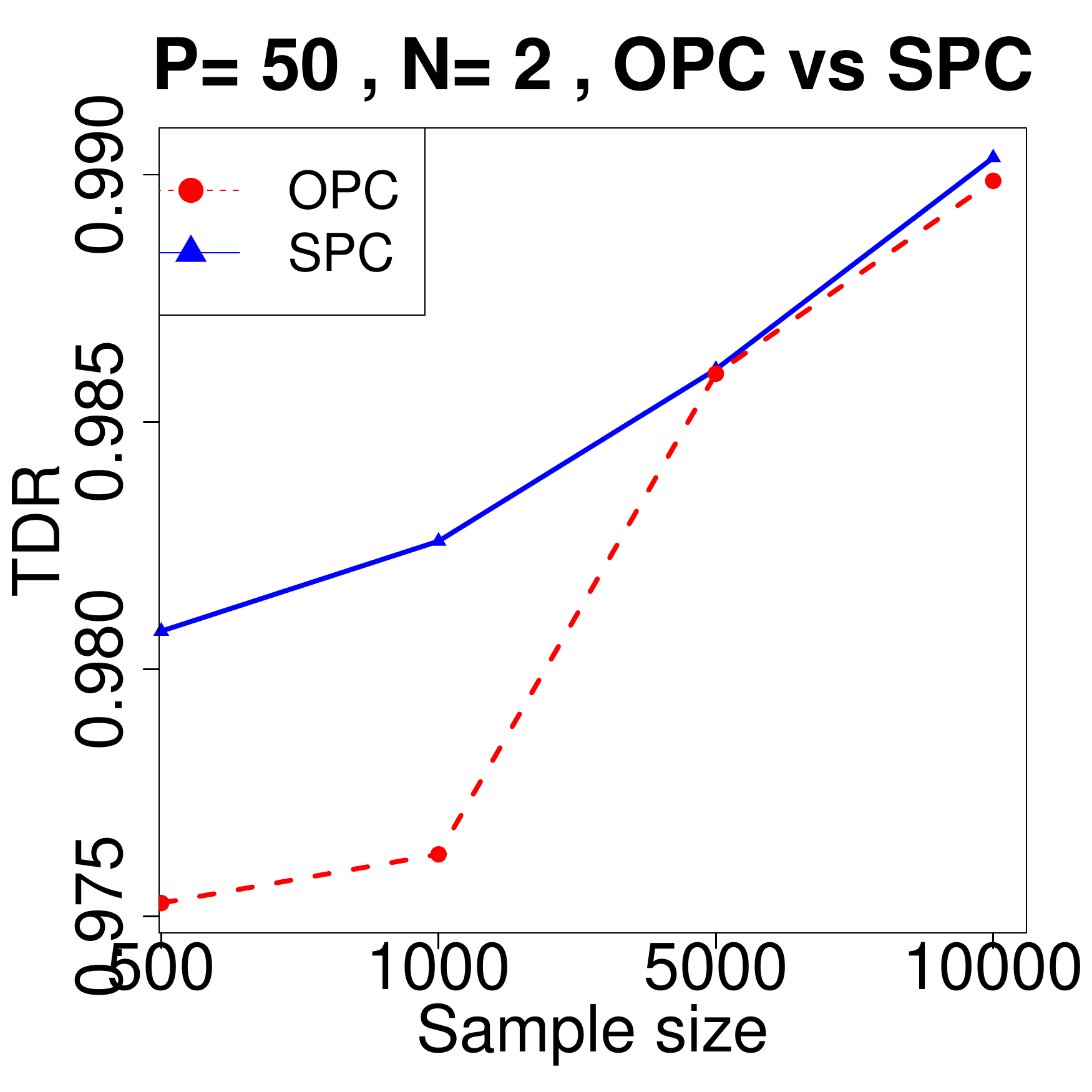}
	\includegraphics[scale=.22,page=3]{results_line_graphs.pdf}
	\includegraphics[scale=.22,page=5]{results_line_graphs.pdf}
	\includegraphics[scale=.22,page=9]{results_line_graphs.pdf}
	\includegraphics[scale=.22,page=7]{results_line_graphs.pdf}
	\includegraphics[scale=.22,page=11]{results_line_graphs.pdf}
	\includegraphics[scale=.22,page=2]{results_line_graphs.pdf}
	\includegraphics[scale=.22,page=4]{results_line_graphs.pdf}
	\includegraphics[scale=.22,page=6]{results_line_graphs.pdf}
	\includegraphics[scale=.22,page=10]{results_line_graphs.pdf}
	\includegraphics[scale=.22,page=13]{results_line_graphs.pdf}
	\includegraphics[scale=.22,page=12]{results_line_graphs.pdf}
	\caption{The first two rows show the performance of the original (OPC) and stable PC-like (SPC) algorithms for randomly generated Gaussian chain graph models:
		average over 30 repetitions with 50 variables  correspond to N = 2, and the significance level $\alpha=0.001$.   The last two rows show the performance of the original (OPC) and stable PC-like (SPC) algorithms for randomly generated Gaussian chain graph models:
		average over 30 repetitions with 1000 variables  correspond to N = 2, sample size S=50, and the significance level $\alpha=0.05,0.01,0.005,0.001$.}
	\label{fig:tprtdrfpr}
\end{figure}

Figure \ref{fig:tprtdrfpr} shows that: (a) as we expected \cite{mxg,Kalisch07}, all algorithms work well on sparse graphs $(N = 2)$, (b) for all algorithms, typically the TPR, TDR, and ACC increase with sample size, (c) for all algorithms, typically the SHD and FPR decrease with sample size, (d) a large significance level $(\alpha=0.05)$ typically yields large
TPR, FPR, and SHD, (e) while the stable PC-like algorithm has a better TDR and FPR in comparison with the original PC-like algorithm, the original PC-like algorithm has a better TPR (as observed in the case of DAGs~\cite{Colombo2014}). This can be explained by the fact that the stable PC-like algorithm tends to perform more tests than the original PC-like algorithm, and (h)  while the original PC-like algorithm has a (slightly) better SHD in comparison with the stable PC-like algorithm in low-dimensional data, the stable PC-like algorithm has a better SHD  in high-dimensional data. Also, (very) small variances indicate that the order-independent versions of the PC-like algorithm in high-dimensional data are stable.
When considering average running times versus sample sizes, as shown in Figure \ref{fig:tprtdrfpr}, we observe that: (a) the average run time increases when sample size increases; (b) generally, the average run time for the original PC-like algorithm is (slightly) better than that for the stable PC-like algorithm in both low and high dimensional settings.

In summary, empirical simulations show that our algorithms achieve competitive results with the original PC-like learning algorithm; in particular, in the Gaussian case the order-independent algorithms achieve output of better quality than the original PC-like algorithm, especially in high-dimensional settings. Algorithm \ref{alg:MVRoriginalPC} and the stable PC-like algorithms have been implemented in the R language (\url{https://github.com/majavid/SUM2019}). 

\bibliographystyle{splncs04}
\bibliography{references}
%

 \section*{Appendix: Proofs of Theorems in Section 4.}
 	Two vertices $x$ and $y$ in  chain graph $G$ are said to be collider connected if there is a chain from $x$ to $y$ in $G$ on which every non-endpoint vertex is a collider; such a chain is called a collider chain. Note that a single edge trivially forms a collider chain (path), so if $x$ and $y$ are adjacent in a chain graph then they are collider connected. The augmented graph derived from $G$, denoted $(G)^a$, is an undirected graph with the same vertex set as $G$ such that $$c\erelbar{00}d \textrm{ in } (G)^a \Leftrightarrow c \textrm{ and } d \textrm{ are collider connected in } G.$$

 	Disjoint sets $X, Y\ne \emptyset,$ and $Z$ ($Z$ may be empty) are said to be
 	$m^\ast$-separated if $X$ and $Y$ are separated by $Z$ in $(G_{an(X\cup Y\cup Z)})^a$. Otherwise $X$ and $Y$ are said to be $m^\ast$-connected
 	given $Z$. The resulting independence model is denoted by $\Im_{m^\ast}(G)$.
 According to \cite[Theorem 3.18.]{rs} and \cite{jv1}, for MVR chain graph $G$ we have: $\Im_m(G)=\Im_{m^\ast}(G)$.

   \noindent\textbf{Theorem 1.}  Let the distribution of $V$ be faithful to an MVR CG $G$, and assume that we are given perfect conditional independence information about all pairs of variables $(u,v)$ in $V$ given subsets $S\subseteq V\setminus \{u,v\}$. Then the output of the stable PC-like algorithm is an MVR CG that has exactly the minimum set of bidirected edges for its equivalence class.

 \begin{proof}
     The proof of Theorem \ref{thm:correctPCstableMVR} is completely analogous to the proof of Theorem 3 and 4 for the original PC-like algorithm in \cite{sp}.
 \end{proof}

   \noindent\textbf{Theorem 2.} The skeleton resulting from the sample version of the stable PC-like algorithm is order-independent.

 \begin{proof}
 We consider the removal or retention of an arbitrary edge $u\erelbar{00} v$ at some level $i$.
 The ordering of the variables determines the order in which the edges (line 7 of Algorithm \ref{alg:MVRstablePC}) and the subsets $S$ of $a_H(u)$ and $a_H(v)$ (line 8 of Algorithm \ref{alg:MVRstablePC}) are considered. By
 construction, however, the order in which edges are considered does not affect the sets $a_H(u)$ and $a_H(v)$.

 If there is at least one subset $S$ of $a_H(u)$ or $a_H(v)$ such that $u\perp\!\!\!\perp_p v|S$, then any
 ordering of the variables will find a separating set for $u$ and $v$. (Different orderings
 may lead to different separating sets as illustrated in Example \ref{ex2OrderDepMVR}, but all edges that have a separating set will eventually be removed, regardless of the ordering). Conversely, if there is no subset $S'$ of $a_H(u)$ or $a_H(v)$ such that $u\perp\!\!\!\perp_p v|S'$, then no ordering will find a separating set.

 Hence, any ordering of the variables leads to the same edge deletions, and therefore to
 the same skeleton.
 \end{proof}

 \noindent\textbf{Theorem 3.}
     Let the distribution of $V$ be faithful to an MVR CG $G$, and assume that we are given perfect conditional independence information about all pairs of variables $(u,v)$ in $V$ given subsets $S\subseteq V\setminus \{u,v\}$. Then the output of the (stable) CPC/MPC-like algorithm is an MVR CG that is Markov equivalent with $G$ that has exactly the minimum set of bidirected edges for its equivalence class.

 \begin{proof}
     The skeleton of the learned CG is correct by Theorem \ref{thm:correctPCstableMVR}.
     Now, we prove that for any unshielded triple $(X_i, X_j, X_k)$ in an MVR CG $G$, $X_j$ is either in all sets that \textit{m}-separate $X_i$ and $X_k$ or in none of them. Since $X_i, X_k$ are not adjacent and any MVR chain graph is a maximal ancestral graph \cite{jv1}, they are \textit{m}-separated given some subset $S\setminus\{X_i, X_k\}$ due to the maximal property. Based on the pathwise \textit{m}-separation criterion for MVR CGs (see section \ref{definitions}), $X_j$ is a collider node in $G$ if and only if $X_j\not\in An(S).$ So, $X_j\not\in S.$ On the other hand, if $X_j$ is a non-collider node then $X_j\in S$, for all $S$ that \textit{m}-separate $X_i$ and $X_k$. Because in this case, $X_j\in An(X_i\cup X_k\cup S)$ and so there is an undirected path $X_i\erelbar{00} X_j\erelbar{00} X_k$ in $(G_{An(X_i\cup X_k\cup S)})^a$. Any set $S\setminus\{X_i, X_k\}$ that does not contain $X_j$ will fail to \textit{m}-separate $X_i$ and $X_k$ because of this undirected path. 
     As a result, unshielded triples are all unambiguous. Since all unshielded triples are unambiguous, the orientation rules are as in the (stable) PC-like algorithm. Therefore,  the output of the (stable) CPC/MPC-like algorithm is an MVR CG that is Markov equivalent with $G$ that has exactly the minimum set of bidirected edges for its equivalence class, and soundness and completeness of these rules follows from Sonntag and Pe\~na \cite{sp}.
 \end{proof}

 \noindent\textbf{Theorem 4.}
     The decisions about \textit{v}-structures in the sample version of the stable CPC/MPC-like algorithm is order-independent.

 \begin{proof}
 The stable CPC/MPC-like algorithm have order-independent skeleton, by
 Theorem \ref{thm:stableskeletonMVR}. In particular, this means that their unshielded triples and adjacency sets are order-independent. The decision about whether an unshielded triple is unambiguous
 and/or a \textit{v}-structure is based on the adjacency sets of nodes in the triple, which are order independent.
 \end{proof}

 \noindent\textbf{Theorem 5.}
     Let the distribution of $V$ be faithful to an MVR CG $G$, and assume that we are given perfect conditional independence information about all pairs of variables $(u,v)$ in $V$ given subsets $S\subseteq V\setminus \{u,v\}$. Then the output of the (stable) LCPC/LMPC-like algorithm is an MVR CG that is Markov equivalent with $G$ that has exactly the minimum set of bidirected edges for its equivalence class.

 \begin{proof}
 By Theorem \ref{thm:correctCPCstableMVR}, we know that the (stable) CPC-like
 and (stable) MPC-like algorithms are
 correct. With perfect conditional independence information, there are no conflicts between orientation rules in the essential graph recovery phase of the
 algorithms. Therefore, the (stable) LCPC-like
 and (stable) LMPC-like algorithms are identical
 to the (stable) CPC-like
 and (stable) MPC-like algorithms.
 \end{proof}

 \noindent\textbf{Theorem 6.}
     The sample versions of (stable) CPC-like
 and (stable) MPC-like algorithms are fully order-independent.

 \begin{proof}
 This follows straightforwardly from Theorems \ref{thm:stableskeletonMVR} and \ref{thm:stablevstructs} and the procedure with lists
 and bi-directed edges discussed above.
 \end{proof}

\end{document}